\documentclass[12pt]{article}
\usepackage{fullpage}
\usepackage{graphicx}
\usepackage{float}
\usepackage{amsmath}
\usepackage{amssymb}
\usepackage{amsthm}
\usepackage{url}
\usepackage{threeparttable}
\usepackage[hmargin=1.0in,vmargin=1.0in]{geometry}
\usepackage{authblk}

\usepackage{color}

\usepackage[markup=underlined]{changes}
\usepackage{todonotes}

\usepackage{fancyhdr}
\usepackage{latexsym}
\usepackage{verbatim}
\usepackage{multirow}
\usepackage{caption} 
\usepackage{natbib}
\usepackage{listings}
\usepackage{titlesec}
\titleformat*{\section}{\LARGE\bfseries}
\titleformat*{\subsection}{\Large\bfseries}
\titleformat*{\subsubsection}{\large\bfseries}

\usepackage{lineno}
\usepackage{setspace}

\usepackage{hyperref}

\begin{document}
\title{Technical note: ShinyAnimalCV: open-source cloud-based web application for object detection, segmentation, and three-dimensional visualization of animals using computer vision}

\author[1]{Jin Wang}
\author[1]{Yu Hu}
\author[2]{Lirong Xiang}
\author[3]{Gota Morota}
\author[1]{Samantha A. Brooks}
\author[1]{Carissa L. Wickens}
\author[1]{Emily K. Miller-Cushon}
\author[1*]{Haipeng Yu}
\affil[1]{Department of Animal Sciences, University of Florida, Gainesville, FL, USA 32611}
\affil[2]{Department of Biological and Agricultural Engineering, North Carolina State University, Raleigh, NC, USA 27695}
\affil[3]{School of Animal Sciences, Virginia Polytechnic Institute and State University, Blacksburg, VA 24061, USA}

\date{}

\maketitle

\noindent 
$^{*}$ Corresponding author: \href{mailto:haipengyu@ufl.edu}{haipengyu@ufl.edu}\\ 

\noindent
ORCID: 0009-0001-4389-4396 (JW), 0009-0008-7061-600X (YH), 0000-0002-3567-6911 (GM), 0000-0002-8923-9733 (HY)\\

\newpage
\doublespacing
\section*{Lay Summary}
The integration of cameras and data science has great potential to revolutionize livestock production systems, making them more efficient and sustainable by replacing human-based management with real-time individualized animal care. However, applying these digital tools to animal data presents challenges that require computer programming and data analysis skills, as well as access to computing resources. Additionally, there is a growing need to train animal science students to analyze image or video data using data science algorithms. However, teaching computer programming to all types of students from the ground up can prove complicated and challenging. Therefore, the objective of this study was to develop ShinyAnimalCV, a user-friendly online web application that allows users to apply data science to analyze animal digital video data. The application includes nine pre-trained models for detecting and segmenting animals in image data and can be easily accessed through a web browser without requiring users to be familiar with programming languages. We have also made the source code and detailed documentation available online for advanced users who wish to use the application locally. This software tool facilitates the application of digital animal data analysis in the animal science community, with potential benefits to research and teaching.\\
\textbf{Teaser Text:} This paper introduces ShinyAnimalCV, an innovative and user-friendly online web application designed for performing computer vision tasks on animal image data. ShinyAnimalCV provides valuable support and enhancements to computer vision research and teaching in the animal science community.

\newpage
\doublespacing
\section*{Abstract}

Computer vision (CV), a non-intrusive and cost-effective technology, has furthered the development of precision livestock farming by enabling optimized decision-making through timely and individualized animal care. The availability of affordable two- and three-dimensional camera sensors, combined with various machine learning and deep learning algorithms, has provided a valuable opportunity to improve livestock production systems. However, despite the availability of various CV tools in the public domain, applying these tools to animal data can be challenging, often requiring users to have programming and data analysis skills, as well as access to computing resources. Moreover, the rapid expansion of precision livestock farming is creating a growing need to educate and train animal science students in CV. This presents educators with the challenge of efficiently demonstrating the complex algorithms involved in CV. Thus, the objective of this study was to develop ShinyAnimalCV, an open-source cloud-based web application. This application provides a user-friendly interface for performing CV tasks, including object segmentation, detection, three-dimensional surface visualization, and extraction of two- and three-dimensional morphological features. Nine pre-trained CV models using top-view animal data are included in the application. ShinyAnimalCV has been deployed online using cloud computing platforms. The source code of ShinyAnimalCV is available on GitHub, along with detailed documentation on training CV models using custom data and deploying ShinyAnimalCV locally to allow users to fully leverage the capabilities of the application. ShinyAnimalCV can contribute to CV research and teaching in the animal science community.

\newpage
\noindent
\textbf{Key words:} computer vision, morphological features, object detection, object segmentation, shiny application, three-dimensional visualization \\
\textbf{Abbreviations:} 2D, two-dimensional; 3D, three-dimensional; AP, average precision; CV, computer vision; CPUs, central processing units; GPUs, graphics processing units; GUI, graphical user interface; PPM, pixels per meter; RGB, red, green, and blue; ROI, region of interest

\newpage
\section*{Introduction}
Precision livestock farming improves current livestock farming systems with an integrated approach based on real-time data collected from sensors and information technology to provide individualized animal care in a timely manner \citep{morota2018big,garcia2020systematic}. Computer vision (CV), a non-intrusive and cost-effective real-time sensor technology, has emerged as a promising  technology to support optimized decision-making in precision livestock farming. 
The availability of affordable two-dimensional (2D) and three-dimensional (3D) camera sensors combined with linear models, machine learning, or deep learning algorithms has facilitated the implementation of CV in livestock production systems. For instance, previous studies have applied CV to monitor animal growth by predicting growth and body composition traits \citep{doeschl2004using,cang2019intelligent,miller2019using,yu2021forecasting}. Other applications of CV include animal identification \citep{andrew2017visual,parham2018animal}, tracking \citep{ahrendt2011development,chen2022vtag}, behavior recognition \citep{nasirahmadi2017new,yang2018feeding,zhang2019real,tsai2020assessment}, and posture inference \citep{zheng2018automatic,riekert2020automatically}. 

Typically, image or video data collected by 2D or 3D camera sensors are first processed to extract morphological features before being fed into various statistical models, machine learning, or deep learning algorithms to accomplish specific computer vision tasks \citep{wang2021asas,li2022barriers}. Despite the availability of computer vision tools, utilizing them often involves complicated processes, such as configuring working environments, pre-processing and post-processing data, training and evaluating models, and visualizing results. Users typically need a certain level of programming and data analysis experience to use these computer vision tools. In addition, computer vision tasks often require substantial computing resources, such as central processing units (CPUs) and graphics processing units (GPUs), due to the computationally intensive nature of the algorithms involved. These factors have limited the widespread adoption of CV systems for research in the animal science community.

With the rapid growth of precision livestock farming, there is a pressing need to train and educate animal science students in CV. However, educators face challenges in effectively explaining the intricate algorithms underlying CV systems to students in the classroom due to their inherent complexity. One viable approach to engage animal science students in CV is to utilize open-source software, which can stimulate a more direct, intuitive, and interactive learning experience \citep{sigut2020opencv}. Unfortunately, there is a lack of relevant CV software to support teaching in the field of animal science.  Fostering the learning of CV among animal science students requires user-friendly open-source software that allows students to execute and understand complex algorithms introduced in the classroom.

Therefore, we developed an open-source cloud-based web application named ShinyAnimalCV. The software holds the potential to support and advance CV research and teaching in the animal science community by providing a user-friendly interface for object detection, segmentation, 3D surface visualization, and 2D and 3D morphological feature extraction.

\newpage
\section*{Software Description}
\subsection*{Overview of software architecture}

\begin{figure}[H]
    \centering  
    \includegraphics[width=\linewidth]{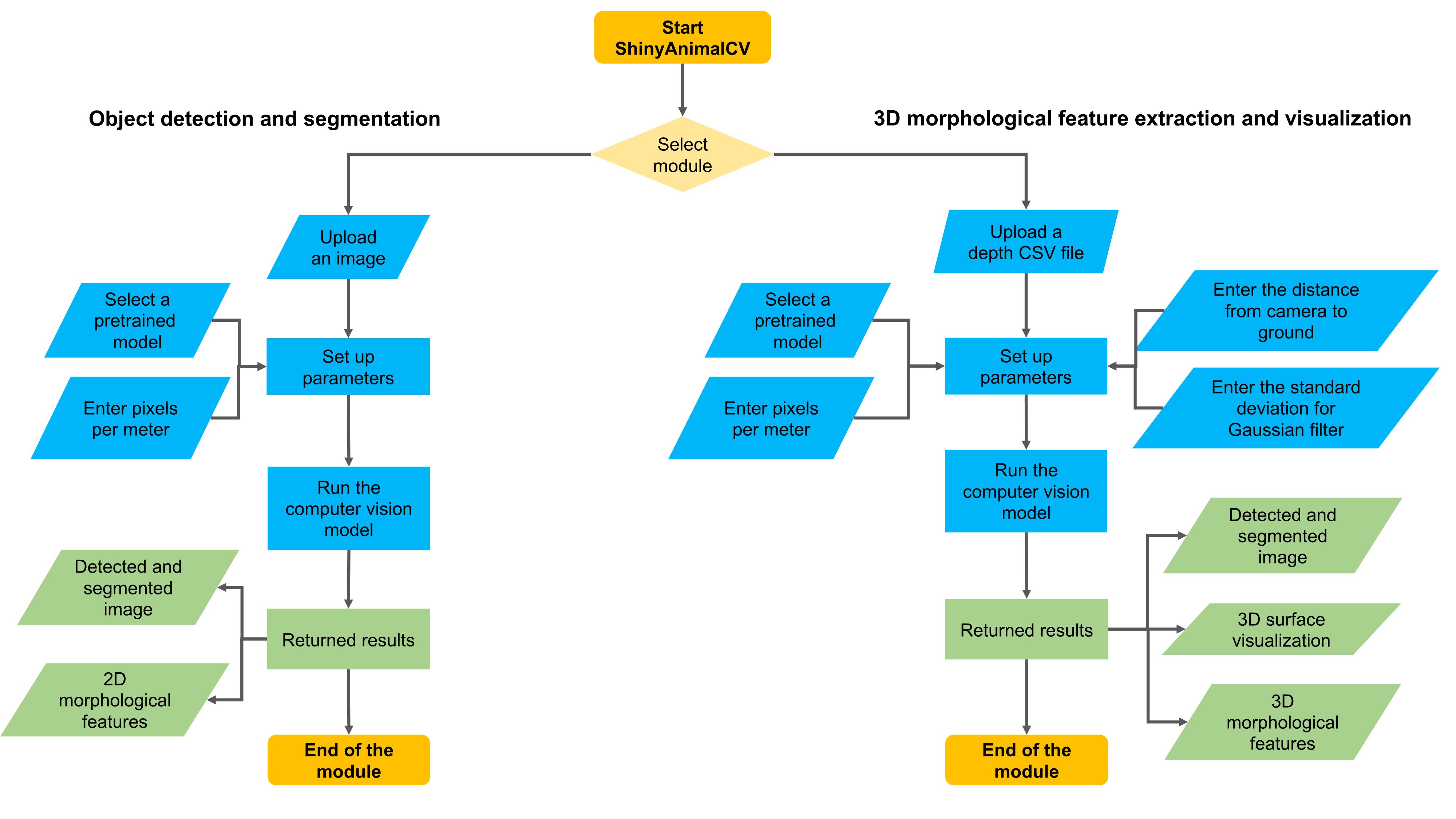}
    \caption{Overall workflow of the two modules in ShinyAnimalCV: object detection and segmentation, and 3D morphological feature extraction and visualization. The start and end points of the modules are highlighted by yellow shapes. User inputs and returned results are indicated by blue and green shapes, respectively.} 
    \label{overalldiagram}
\end{figure}

The presented ShinyAnimalCV is built by the integration of R \citep{R} and Python \citep{Python} using the reticulate package \citep{Reticulate}. The application consists of two components: a web-based graphical user interface (GUI) and a server. The GUI was developed using the shiny R package \citep{shiny}, which provides a user-friendly interface. The GUI page consists of an information page and two computer vision modules: object detection and segmentation, and 3D morphological feature extraction and visualization. The information page includes a link to the documentation on GitHub with detailed instructions and video tutorials covering the use of the two modules. The overall workflow of the two modules is described in Figure \ref{overalldiagram}. In brief, the GUI allows the user to import data (e.g., images) and set up model parameters. The collected user input is then passed to the server side for analysis, and the output is returned to the GUI. The implementation of shiny reactive expressions in ShinyAnimalCV dynamically updates user input and uses resources efficiently to reduce the computational burden. 

We implemented ShinyAnimalCV using a modified version of the state-of-the-art python-based computer vision model Mask R-CNN \citep{he2017mask} on the server side to perform CV tasks along with other Python libraries, including matplotlib \citep{Matplotlib}, OpenCV \citep{opencv_library}, and plotly \citep{plotlyR}. Leveraging the shiny R package, these computer vision models and packages were seamlessly encapsulated and transformed into a web application, providing a user-friendly interface that can be operated using only a mouse within a web browser. This greatly improves the accessibility of the software, particularly for users with limited programming experience and access to computing resources. ShinyAnimalCV is deployed online using an instance on the supercomputer (HiPerGator) at the University of Florida, with 32 Intel(R) Xeon(R) CPU E5-2683 v4 @ 2.10GHz processors, 8 NVIDIA GeForce GTX 1080 Ti GPUs, and 125 GB of random access memory, and is accessible at \href{https://shinyanimalcv.rc.ufl.edu/}{https://shinyanimalcv.rc.ufl.edu/}.

\subsection*{Object detection and segmentation}
\begin{figure}[H]
    \centering  
    \includegraphics[width=\linewidth]{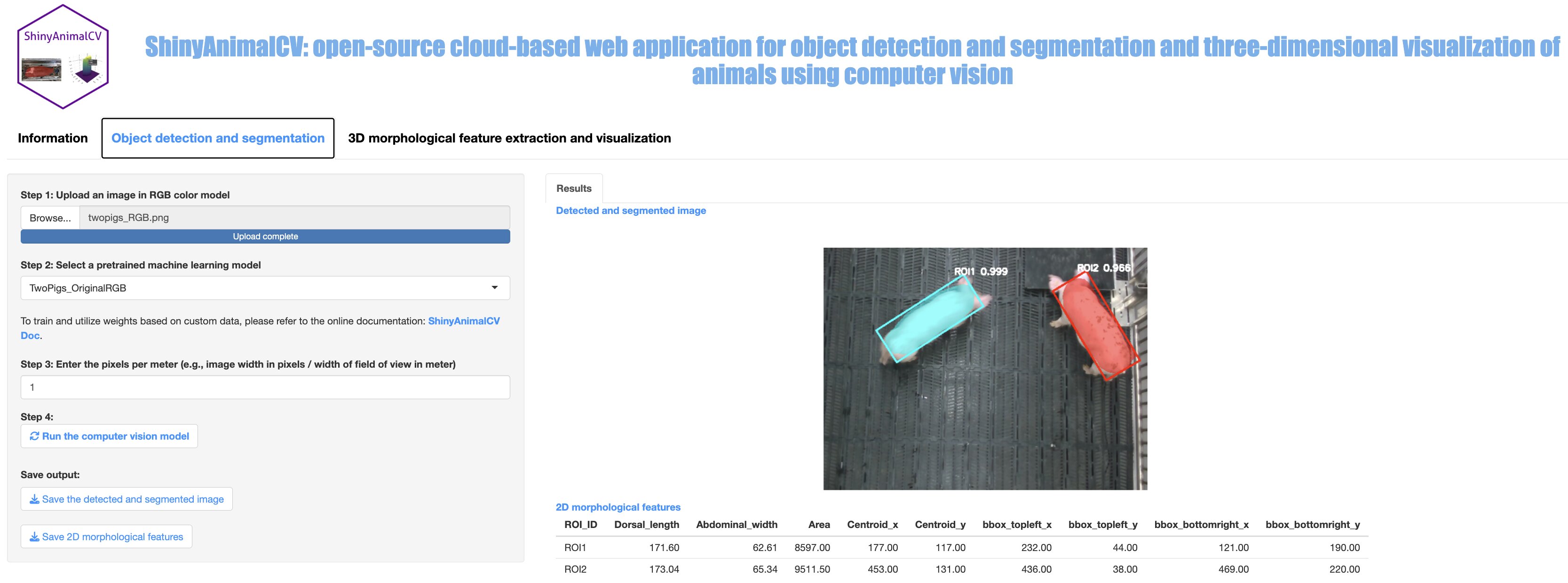}
    \caption{The interface of the object detection and segmentation module in ShinyAnimalCV. On the left panel, users can upload the input RGB image and configure various parameters, such as selecting a pre-trained model and setting the pixels per meter value. Additionally, users have the option to download the results. The right panel displays the returned results, which include the detected and segmented image along with the 2D morphological features.} 
    \label{objmodule}
\end{figure}

The object detection and segmentation module allows users to perform object detection and instance segmentation of animals using top-view images, while also extracting 2D morphological features from the detected and segmented results (Figure \ref{objmodule}). The workflow consists of 4 steps. In step 1, the file input button allows users to upload a single red, green, and blue (RGB) or depth image at a time to perform object detection and segmentation. These models have been trained on RGB color model by treating the depth image as RGB in PNG format, so input images with RGB color model are required to obtain accurate results for object detection and segmentation. However, various image formats, such as JPG and TIFF, are supported. Step 2 provides a select input button with a set of machine learning models pre-trained on top-view animal image data. In step 3, the numeric input button allows users to enter the pixels per meter (PPM) value based on the input image. This value can be then used to convert the extracted morphological features, measured in pixels, to a meter-based unit. The default value for PPM is set to 1. After entering all required inputs, users can perform object detection and segmentation by clicking the action button in step 4.

The returned results include a detected and segmented image, along with a table displaying the extracted 2D morphological features (Figure \ref{objmodule}). Each detected instance is labeled with a unique color (i.e., mask) to highlight the region of interest (ROI). A modified minimum rotated rectangle bounding box based on the labeled/segmented mask is created using OpenCV to localize and enclose the detected object. This modified bounding box is designed to replace the original axis-aligned rectangle bounding box from Mask R-CNN, resulting in more precise 2D morphological features. A confidence score ranging from 0 to 1 is provided to indicate the model's confidence or probability regarding the detection accuracy. A set of 2D morphological features are extracted based on the detection and segmentation results and presented in a table, including dorsal length, abdominal width, total area, centroid coordinates, and bounding box coordinates (Figure \ref{objmodule}). Detailed descriptions of these features can be found in Table \ref{MTdef}. By default, the unit of morphological features is based on pixels (PPM = 1). The Save output section provides download buttons that allow users to locally save the detection and segmentation image as a PNG file, and morphological features as a CSV file for  convenient access and subsequent analysis. 

\subsection*{3D morphological feature extraction and visualization}
This module processes depth map files obtained from depth sensors (e.g., Intel RealSense D400 series) for a single animal to extract 3D morphological features and provide interactive visualization of the depth data. The depth map files record the distance from the camera to the surface of objects. The workflow consists of 6 steps. Step 1 provides a file input button to collect the user-provided depth map file in CSV format. In step 2, users can choose a CV model from a list of provided pre-trained models. Steps 3 and 4 allow users to enter a PPM and a distance from the camera to the background (2.5 meters by default) using a numeric input button, respectively. In step 5, users have the option to input a value for the standard deviation of the Gaussian filter, which determines the level of smoothing applied to the 3D surface. Once all inputs are entered, the module can be executed by clicking the action button. In detail, the uploaded depth map file from step 1 is first converted to an RGB heatmap image using the matplotlib package. The Mask R-CNN model with the pre-trained model weight selected in step 2 is used to segment and label the ROI in the converted image, resulting in a binary mask. The binary mask contains pixels with values of 0 or 1, where a value of 1 represents the ROI and 0 corresponds to the background. The uploaded depth file from step 1 is then cleaned to retain only the ROI by applying a Hadamard product between the depth file and the binary mask obtained in the previous step. The cleaned depth file is further refined by replacing all zeros (missing) with the mean value of the cleaned depth file (excluding zeros), followed by filtering the noise using Gaussian filtering. The resulting cleaned and filtered data is then employed to extract 3D morphological features and visualize them interactively using the plotly package.

\begin{figure}[H]
    \centering  
    \includegraphics[width=\linewidth]{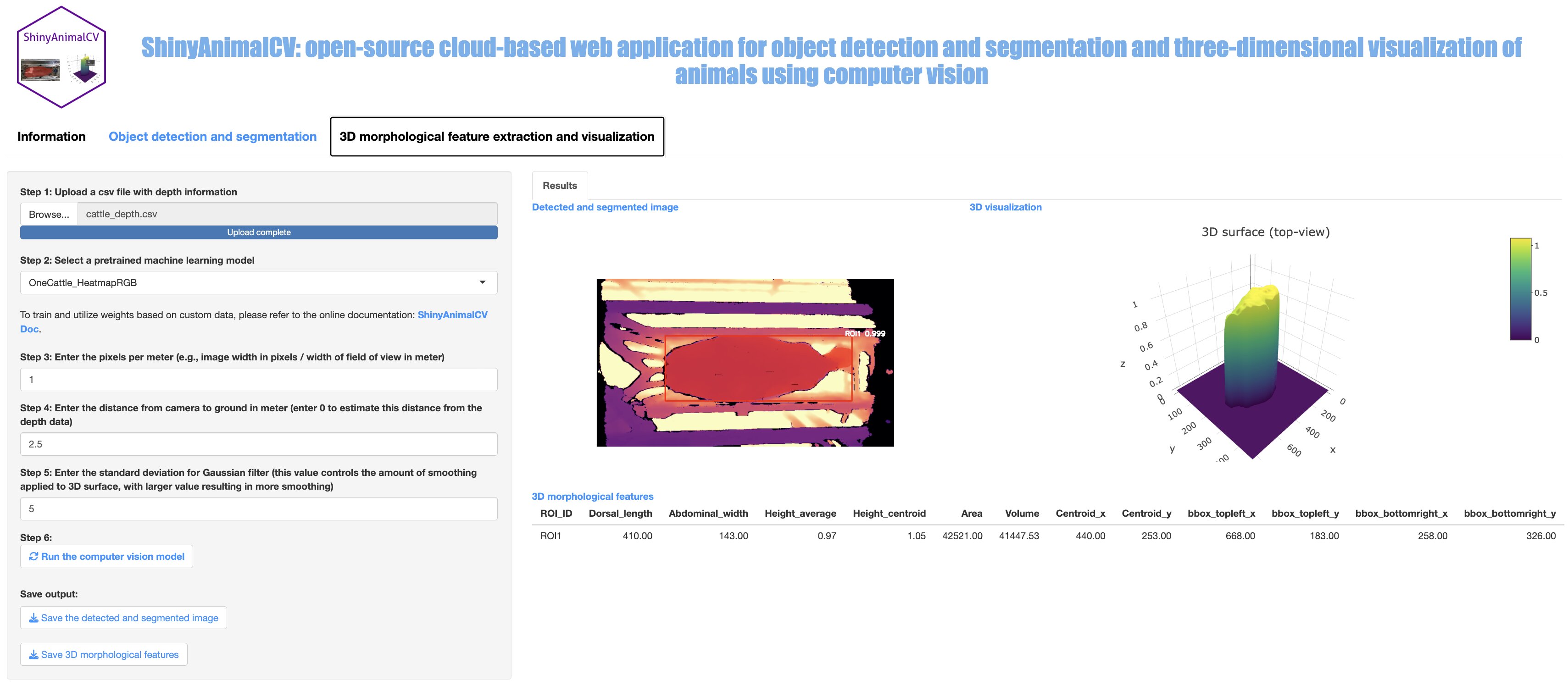}
    \caption{The module of 3D morphological feature extraction and visualization in ShinyAnimalCV. The left panel allows users to upload the input depth file in CSV format and set up parameters including a pre-trained model, pixels per meter, distance from the camera to ground in meters, and the standard deviation of the Gaussian filter. The right panel displays the results of the detected and segmented image, as well as 3D visualization and morphological features. Users can download these results using the Save output section on the left panel.} 
    \label{3dmodule}
\end{figure}

The results returned from this module include a detected RGB heatmap image, an interactive 3D plot, and a table of 3D morphological features (Figure \ref{3dmodule}). The detected RGB heatmap image is generated using the object detection and segmentation module, which returns a detected ROI accompanied by a label, confidence score, colored predicted mask, and minimum rotated bounding box. The 3D plot created with the cleaned depth map file allows users to explore and visualize the depth data in a dynamic and engaging manner (e.g., zooming, rotating, and interacting). The extracted 3D morphological features consist of all 2D features described in the object detection and segmentation module and additional features, including average height, the height of the detected object centroid (geometric center), and total volume. The detailed descriptions of these features are shown in Table \ref{MTdef}. The detected and segmented RGB heatmap image and the table of extracted 3D morphological features can be easily downloaded and accessed locally using the provided download buttons in the Save output section.   

\subsection*{Pre-trained computer vision models}
Currently, the nine pre-trained models included in ShinyAnimalCV support top-view pig and dairy cattle image data. Images with an RGB color model in PNG format were used to train the models, including single, two, and four pigs and single dairy cattle. The total number of training images was 120 for single pig and dairy cattle, and 200 for two and four pigs, respectively. The performances of all trained models were evaluated using the average precision (AP) metric at two thresholds of intersection over union (IoU): 0.5 and 0.75. The AP is calculated by averaging the precision over all recall values from $0$ to $1$, which is defined as $$ \int_{0}^{1} p(r) dr,$$ where $p(r)$ refers to the precision at recall $r$. Table \ref{maskrcnnperformance} provides detailed information on the model training and evaluation results. The pig and dairy cattle data used for model training were obtained from \cite{yu2021forecasting} and \cite{kadlec2022automated}, respectively.

\subsection*{Software limitations and proposed solutions}

While the ShinyAnimalCV software includes nine pre-trained models, one limitation is that these models may not always be applicable to users' data in different scenarios or species. To overcome this limitation, we have created a comprehensive tutorial that allows users to annotate or label their own custom data, train CV models using labeled data, and perform inference on their data using trained models. Although ShinyAnimalCV is deployed online with access to significant computing resources, we recognize that some users may prefer to use local computing resources to process a large number of images. To address this need, we have made the source code of ShinyAnimalCV publicly available on GitHub, as well as provided step-by-step instructions for users to install and run ShinyAnimalCV locally. This allows users to take full advantage of the software's capabilities. We believe that these resources, including instructions and tutorials, will empower users to leverage ShinyAnimalCV to its maximum potential. All instructions and tutorials are available online at: 
\href{https://github.com/uf-aiaos/ShinyAnimalCV}{https://github.com/uf-aiaos/ShinyAnimalCV}.

\newpage
\section*{Conclusion and Outlook}

ShinyAnimalCV is an open-source cloud-based web application for object detection, segmentation, 3D visualization, and 2D and 3D morphological feature extraction. We have deployed ShinyAnimalCV available online at \href{https://shinyanimalcv.rc.ufl.edu/}{https://shinyanimalcv.rc.ufl.edu/}, providing a user-friendly GUI for users with limited programming experience and access to computing resources. The source code of ShinyAnimalCV and detailed installation instructions are available on GitHub: \href{https://github.com/uf-aiaos/ShinyAnimalCV}{https://github.com/uf-aiaos/ShinyAnimalCV}. The full transparency of the code allows users to use the software locally and contribute to the development of ShinyAnimalCV. ShinyAnimalCV is anticipated to support and advance CV research and teaching in animal science.

\newpage
\section*{Acknowledgments}
This work was supported by the University of Florida startup funds to H.Y. We thank Oleksandr (Alex) Moskalenko for assisting us in setting up the instance on HiPerGator to deploy ShinyAnimalCV.

\section*{Conflict of Interest Statement}
The authors declare no real or perceived conflicts of interest.

\clearpage
\newpage 
\bibliographystyle{apalike} 
\bibliography{ShinyAnimalCV}

\newpage
\begin{table}[H]
\begin{center}
  \caption{List of 2D and 3D morphological features returned by ShinyAnimalCV. PPM and bbox refer to pixels per meter and minimum rotated bounding box, respectively. The units of the morphological features are defined when PPM is equal to 1. }
  \label{MTdef}
  \scalebox{0.85}{
    \begin{tabular}{lllll}
      \multicolumn{2}{c}{Morphological features} & \multicolumn{2}{c}{Definition of the features} & Unit \\ \hline
      & Abdominal width & width of the bounding box localizing the object &  & pixel \\ 
      & Area  & segmented object's total area, defined as the mask area. & & pixel\textsuperscript{2} \\ 
      & bbox\_topleft\_x  & row coordinate of the top-left corner of the bounding box & & pixel \\
      & bbox\_topleft\_y  & column coordinate of the top-left corner of the bounding box & & pixel \\
      & bbox\_bottomright\_x   & row coordinate of the bottom-right corner of the bounding box & & pixel  \\
      & bbox\_bottomright\_y  & column coordinate of the bottom-right corner of the bounding box & & pixel \\
      & bbox\_rotatedangle  & rotated angle of the bounding box & &   degree\\
      & Centroid\_x  & row coordinate of the center of the segmented object& & pixel \\ 
      & Centroid\_y  &column coordinate of the center of the segmented object& & pixel \\
      & Dorsal length & length of the bounding box localizing the object & & pixel\\ 
      & Height\_average & average height over the region of the segmented object  & & meter\\
      & Height\_centroid &  height at the center of the segmented object  & & meter\\
      & Volume & sum of heights over the segmented object & & meter\textsuperscript{3}\\
      
      \hline
      \hline
    \end{tabular}
  }
\end{center}
\end{table}

\newpage
\begin{table}[H]
\begin{center}
  \caption{The performance of nine pre-trained models is evaluated based on the average precision (AP) at the intersection over union 0.5 and 0.75. These models were trained using top-view pig and dairy cattle RGB, depth (Dep), and depth map file converted heatmap (Hm) images. For training the single pig and cattle models, 120 images were used, while two and four pigs models were trained with 200 images. The performance of all models was evaluated using 40 images.}
  \label{maskrcnnperformance}
  \scalebox{0.93}{
    \begin{tabular}{cccccccccccccc}
      \hline \hline
      & & \multicolumn{8}{c}{Top-view pig} &  \\
      \cline{3-10}
      & & \multicolumn{3}{c}{Single} & & \multicolumn{2}{c}{Two} & & Four & & \multicolumn{3}{c}{Top-view cattle} \\ 
      \cline{3-5}
      \cline{7-8}
      \cline{10-10}
      \cline{12-14}
      & Metrics & RGB & Dep & Hm & & RGB & Dep & & RGB & & RGB & Dep & Hm\\ 
      \hline
      & $AP_{0.5}$ & 1.0 & 1.0 & 1.0 & & 0.95 & 1.0 & & 0.99 & & 1.0 & 1.0 & 1.0 \\ 
      & $AP_{0.75}$ & 1.0 & 0.85 & 0.85 & & 0.84  & 0.89 & & 0.93 & & 1.0 & 1.0 & 1.0 \\ 
      \hline
    
      \hline
    \end{tabular}
  }
\end{center}
\end{table}

\newpage 

\newpage 

\newpage 

\end{document}